# *SANE*: *S*trategic *A*utonomous *N*on-Smooth *E*xploration for Multiple Optima Discovery in Multi-modal and Non-differentiable Black-box Functions

Arpan Biswas[1,a], Rama Vasudevan[2], Rohit Pant[3], Ichiro Takeuchi[3], Hiroshi Funakubo[4], Yongtao Liu[2,b]

[1] University of Tennessee-Oak Ridge Innovation Institute, Knoxville, TN 37996, USA

[2] Center for Nanophase Materials Sciences, Oak Ridge National Laboratory, Oak Ridge, TN 37830, USA

[3] Department of Materials Science and Engineering, University of Maryland, College Park, Maryland 20742, United States of America

[4] Department of Materials Science and Engineering, Tokyo Institute of Technology, Yokohama, 226-8502, Japan

## Abstract

Both computational and experimental material discovery bring forth the challenge of exploring multidimensional and multimodal parameter spaces, such as phase diagrams of Hamiltonians with multiple interactions, composition spaces of combinatorial libraries, material structure image spaces, and molecular embedding spaces. Often these systems are black-box and time-consuming to evaluate, which resulted in strong interest towards active learning methods such as Bayesian optimization (BO). However, these systems are often noisy which make the black box function severely multi-modal and non-differentiable, where a vanilla BO can get overly focused near a single or faux optimum, deviating from the broader goal of scientific discovery. To address these limitations, here we developed Strategic Autonomous Non-Smooth Exploration (SANE) to facilitate an intelligent Bayesian optimized navigation with a proposed cost-driven probabilistic acquisition function to find multiple global and local optimal regions, avoiding the tendency to becoming trapped in a single optimum. To distinguish between a true and false optimal region due to noisy experimental measurements, a human (domain) knowledge driven dynamic surrogate gate is integrated with SANE. We implemented the gate-SANE into a pre-acquired Piezoresponse spectroscopy data of a ferroelectric combinatorial library with high noise levels in specific regions, and a piezoresponse force microscopy (PFM) hyperspectral data. SANE demonstrated better performance than classical BO to facilitate the exploration of multiple optimal regions and thereby prioritized learning with higher coverage of scientific values in autonomous experiments. Our work showcases the potential application of this method to real-world experiment, where such combined strategic and human intervening approaches can be critical to unlocking new discoveries in autonomous research.

## Introduction

Recent advancement of automated and autonomous experiments has been transforming the landscape of scientific research.[1–5] By integrating experiment automation with machine learning-enabled data analysis and decision-making processes, researchers can now conduct experiments at an unprecedented pace, accelerating the discovery of new materials and the process of characterizing and understanding complex materials systems.

Bayesian Optimization (BO)[6–8] is an active learning method which aims to autonomously explore the parameter space and continually learns the unknown ground truth and its global optimal region, where the ground truth function can be either black-box or expensive to evaluate, or both. Given a few evaluated training samples, the expensive unknown function is replaced by a cheaper surrogate model (*e.g.* Gaussian Process)[9–11], and the surrogate model continues to learn the human defined region of interests with adaptive selection (*e.g.* Acquisition function)[12–15] of locations for future expensive evaluations. BO is more popular than other design of experiment methods (*e.g.* Random sampling, Latin Hypercube sampling, *etc.*) as it is designed to converge to the optimality with minimal expensive evaluations. Thus, BO has attracted special attention in the materials science domain where accelerated BO driven autonomous discovery has been particularly impactful, enabling efficient identification of optimal conditions for particular material properties without human intervention, such as bandgap optimization[16], small-molecule emitters discovery, maximizing carbon nanotube growth rates[17], and so on. This type of autonomous workflow with BO has been widely used in recent studies to adaptively explore expensive control parameter spaces of physical/simulation models[18–23] and experiments[24], material structure-property relationship discovery[25] and to develop autonomous platforms towards accelerating chemical[26,27] and material design[28–30]. Recently, interactive BO frameworks through minor human intervention (human in the loop) proved to have better material processing[31] and microscope experimental steering[32–37]. A number of excellent reviews on BO are available[6,38], and it is now implemented in a broad range of Python libraries including BOtorch (Pytorch)[39] and Gpax (Jax)[40].

However, the pursuit of optimal conditions, while valuable for applications-focused materials development, may not always align with the broader goals of scientific discovery. In the context of scientific and knowledge discovery, understanding the relationships between the vast parameter space and resulting material properties is crucial. The challenge arises when the experimental process becomes overly focused on finding a singular "optimal" condition. This focus can limit the exploration of sub-optimal or diverse conditions that can offer deeper insights into the material's behavior, ultimately advancing our understanding of the underlying principles governing material properties. Moreover, real-world experiments are often subject to noise and other uncertainties, which can lead to "fake" optima—conditions that appear to be optimal due to experimental error or noise but do not represent the true optima. Also, due to such noise, the overall parameter space becomes very non-smooth, e.g., introducing multi-modality and high non-differentiability and therefore becoming much harder to explore, learn the parameter space and locate optimal conditions. In the theory of optimization, numerical methods generally struggle with highly non-smooth functions[41] and BO is not an exception due to the surrogate model priors over the belief of a smooth function. It is evident to say that it is even harder to solve a black-box non-smooth function. One can, however, project to a smoother function from such non-smooth functions with the help of either domain knowledge[8] or by fitting a cluster of GPs which learn each localized smooth function.[42] However, it is a challenging ask to possess appropriate domain knowledge for such projections, or know the number of clusters or region of non-smoothness apriori. Also, in the case of structure-property learning, the exploration needs to be done over the raw non-smooth function space rather than a projected space to avoid losing critical insights. Thus, when exploring over a non-smooth function due to noisy experiments, when the decision-making process is based on these potentially misleading optimal conditions, there is a risk of missing out on critical information that lies in other sub or near optimal regions. As a result, the classical BO methods may inadvertently limit the scope of exploration in autonomous experimentation.

To address these limitations, we have developed a BO driven Strategic Autonomous Non-Smooth Exploration (SANE) workflow that is designed to navigate and mitigate the challenges posed by noisy experimental data and the tendency to become trapped in local optima. Our SANE framework emphasizes the exploration of a broader range of conditions, including sub-optimal and diverse points within the parameter space. This approach ensures a more comprehensive understanding of the material system being studied, enabling the discovery of new insights and correlations that may otherwise be overlooked. We implemented this SANE approach in two distinct experimental datasets: (1) a piezoresponse spectroscopy hysteresis loop dataset across composition spread of Sm-doped $BiFeO_3$ with high noise levels in specific regions, and (2) a grid piezoresponse spectroscopy hyperspectral dataset over various domain structures of a $PbTiO_3$ thin film. These two datasets have been explored in our previous studies,[43–45] indicating that they are good model experimental datasets to demonstrate the application of newly developed ML approaches. In both cases, SANE demonstrated its ability to avoid becoming stuck in noisy regions or singular optima and facilitated the exploration of multiple optimal and sub-optimal conditions. This not only improved the robustness of the optimization process but also provided a more complete picture of the parameter space.

We further enhance our SANE method by adapting a recently developed gated active learning approach[45] that allows for the incorporation of human expertise and domain knowledge into the autonomous experimentation process. This hybrid approach leverages the strengths of both autonomous systems and human intuition, allowing for more targeted and effective exploration. By guiding the optimization process with human input constraints, we can prioritize the parameter space that has higher scientific value, thereby improving the overall performance and outcomes of autonomous experiments. Our work showcases the potential of this method through its application to noisy experiment, demonstrating its efficacy in avoiding local optima and uncovering a richer understanding of the material systems under study. As the field of materials science continues to evolve, the development and adoption of such strategic approaches will be critical in unlocking new discoveries and pushing the boundaries of what is possible in autonomous research.

## Results and Discussion

1. **SANE framework**

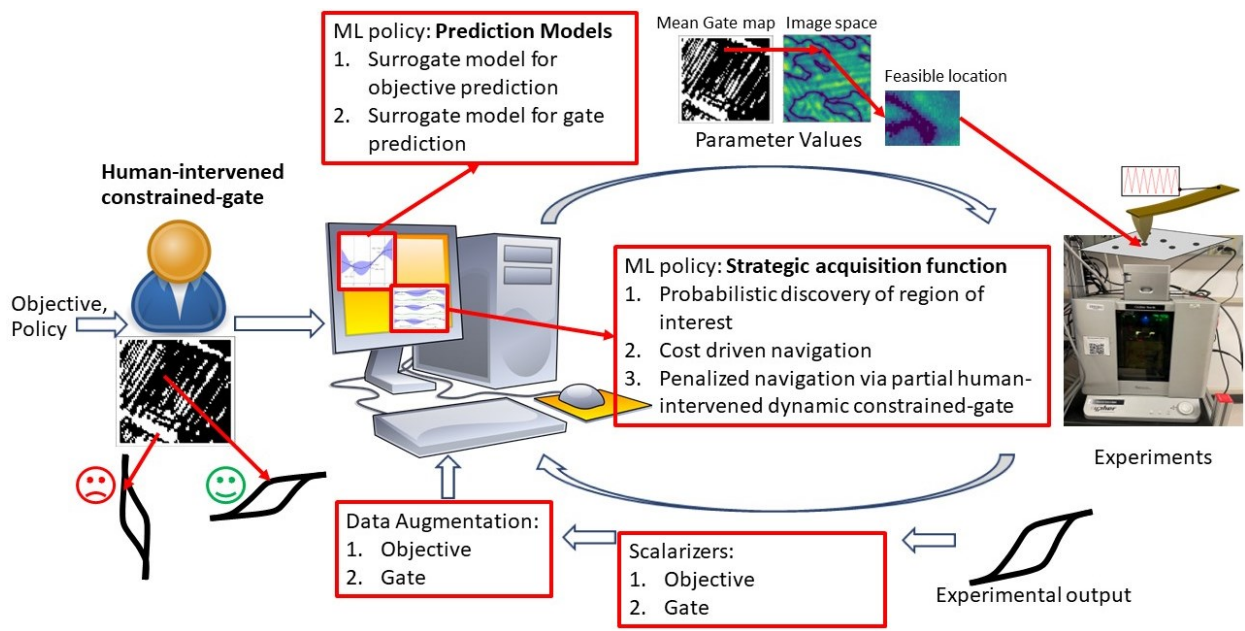

**Figure 1.** Workflow of Strategic Autonomous Non-Smooth Exploration (SANE) with partial human-intervened dynamic constrained gate. Here, the SANE workflow is shown over AFM experiments but can be implemented to any material characterization techniques or any microscopy measurements. Here, the research contribution is the development of the ML policy for a strategic acquisition function, which is subcategorized into a) probabilistic discovery of region of interest, b) cost-driven navigation and c) penalized navigation to explore on feasible search space only, as defined from partial human-intervened dynamic gate. Here the subject matter expert assesses the quality of the experimental result (eg. spectral structure) at the initialization of the SANE, and an initial estimated gate map is defined. Then, the estimated gate map is updated with strategic BO driven autonomously sampled new data, without any human intervention.

**Fig. 1** provides the overall architecture of the proposed SANE workflow with human-intervened dynamic constrained gate, developed over the naïve BO workflow. Unlike in traditional BO, where the acquisition function assumes uniform cost during exploration over the search space, SANE performs a goal orientation exploration with a non-uniform cost of the search space. Cost acquired BO has been developed earlier[46–48], however, the cost function is formulated based on the invariable cost of the experiments over the search space. In this case, we formulate the cost function based on the strategies of 1) discovering a potential global or local region of interest, 2) guiding the exploration centering on new regions of interest and 3) guiding the exploration centering on the previously found region of interests. SANE is initialized as BO where we set the total number of iterations as $N$. $N$ can be defined as the total cost based on a reasonable time the autonomous experimentation can be performed. As the iteration $i$ progress, after every $n \ll N$ iterations, we have a check for discovering a new region of interest. If solution superior to the current focused solution is found within the samples explored between the current check and previous check, a current or potential global region of interest is found. Else, we perform an optimization routine to find a local optimal sampled after the current focused sample, and then a probabilistic check to determine if the local optimal belongs in a potential region of interest. This optimization scheme is formulated as per eqs. 1-5. Once the local optimum is found, the binary check of belonging to a region of interest ($1 - Yes, 0 - No$) is governed by the sampling from logistic distribution with probability, $p_{ROI=1}$ and as $p_{ROI=0}$ respectively per eq. 6. Let's assume, at iteration $i$, we have already determined $k$ focused locations. Here, $f_1$ is defined as the absolute distance in the normalized parameter space between the current focused location $X_{f,k}$ and the location sampled after $X_{f,k}$. $f_2$ is defined as ratio between the output of the location sampled after $X_{f,k}$ and the output $y_{f,k}$ at $X_{f,k}$. Here, $f_3$ is defined as the mean of the absolute distance in the normalized parameter space between each of the previously focused location $X_{f,.}$ and the location sampled after $X_{f,.}$. As a strategic approach to discover the new region of interest, we would prefer to choose the sample with the farthest distance from current and previously all the focused locations, and closest to the output of the currently focused location. Thus, as per the optimization (maximization) routine, the higher values of combined product of $f_1, f_2, f_3$ are chosen. Though it is not always that this best solution belongs to another region of interest, the strategic approach does mean that the likelihood increases with higher values of $f_1, f_2$. Therefore, we make a probabilistic decision whether to navigate to the new focused location via sampling from the logistic distribution with probability of mean value of $f_1, f_2$. With higher probability, we are likely to add and navigate towards the new focused location ($X_{f,k+1}$) and with lower probability, we are likely to consider the current focused location ($X_{f,k+1} = X_{f,k}$) for the upcoming navigation governed by the strategic cost function.

$$f_1 = |X - X_{f,k}| \tag{1}$$

$$f_2 = y/y_{f,k} \tag{2}$$

$$f_3 = \frac{\sum_{j=1}^{k-1}|X - X_{f,j}|}{k-1} \tag{3}$$

$$\max_X F = \frac{f_1 \times f_2 \times f_3}{\dim(X)} \tag{4}$$

$$X \in \{X_{\arg(X_{f,k})}, \ldots \ldots, X_i\} \subseteq D_i \tag{5}$$

$$p_{ROI=1} = \frac{f_1 + f_2}{\dim(X)} \tag{6}$$

$$p_{ROI=0} = 1 - p_{ROI=1} \tag{7}$$

Next, we define the navigation strategy considering all the focused locations. In SANE, strategic exploration is defined as the likelihood of discovering a new region of interest while avoiding becoming trapped over the discovered regions of interest. Whereas strategic exploitation is defined as the likelihood of unearthing critical information over all the discovered regions of interest (optimal and suboptimal points), without only focusing on the singular optimal region. This navigation is governed by a strategic cost function and thus depending on the current state (say iteration $i$), the cost function puts variable weightage of performing exploration or exploitation. The strategic cost-based acquisition function can be described step by step as follows. Given the mean and variance of the unexplored data from the prediction model as fitted from the explored data, the standard acquisition values of all the unexplored locations are calculated as $u(X), X \notin \boldsymbol{D_i}$. Here, we can follow any standard acquisition functions in BO such as Probability of Improvement (POI), Expected Improvement (EI) or Upper or Lower Confidence bound (UCB/LCB). Then, the cost driven acquisition function $u_c(X)$ is can be either exploitative in preference to choose the next sample for function evaluations with the closest distance from current and previously focused locations, or explorative to prefer to choose the same with the farthest from current and previously focused locations. This can be mathematically formulated as per eq. 8, assuming at iteration $i$ and $k$ focused locations. Here, $\boldsymbol{s}$ is a switching parameter trajectory with iterative binary choices between strategic exploitation and exploration. $\boldsymbol{s}$ can be pre-defined from the domain expert preference of navigation in the parameter search space. Additionally, to avoid any navigation instability arising due to negligible standard acquisition values almost over the entire search space, we perform the cost driven acquisition function with fully localized search when $g \geq \alpha$. $\delta = 10^{-5}$ is a very small value for numerical stability and $\alpha$ is a user defined value between 0 and 1, recommended to be anything $\alpha \geq 0.9$. This navigation instability can arise when the standard acquisition function provides false values of negligible information gain in exploring in almost any location during the phase of near optimal learning. However, in a real scenario, this is often not true due to the complex and noisy parameter space and more search is needed and potential better insights can be learned. For strategizing a fully localized search, we would prefer to exploit the current focused locations and avoid exploring again to previously focused locations. This is because such scenarios occur mostly at near optimal learning, where most of the previously focused locations are already exploited and therefore to save cost, we put weightage on the local exploration and exploitation only rather than global exploration and exploitation. Thus, we choose the next best sample which has the minimum distance to the current focused locations and maximum distance to the previously focused locations.

$$u_c(X) = \begin{cases} \frac{u(X)}{f_1 + f_3} & , g_i \leq \alpha, s_i = 0 \\ u(X) \times (f_1 + f_3), & g_i \leq \alpha, s_i = 1 \\ f_3 - f_1 & , g \geq \alpha \end{cases} \quad (8)$$

$$g_i = \frac{\max(u(X)) + \delta}{\sum_{X \notin D_i} u(X) + \delta} \quad (9)$$

Finally, the SANE workflow is further enhanced with a human assessed constrained gate, to avoid exploring and exploiting "fake" regions of interest due to noisy measurements, thereby avoiding gathering false insights. It is to be noted that the distinction between the strategy of earlier mentioned probabilistic finding of region of interest and the human assessed gate is that the constrained gate narrows down the feasible search space, and within the search space SANE aims to discover the region of interest with the probabilistic approach. In other words, without the gate, SANE can guide to focus on a "fake" region of interest with high probability. Without the region of interest finding strategy, the gate itself will only separate the feasible space but not the discovery of the multiple optimal and suboptimal points (hills in case of maximization). Previously, human in the loop based autonomous exploration workflows have been developed which proved to have better performance and alignment with the expectation to the experimentalists, than a purely data driven active learning method.[32,34–37,49,50] To avoid the increment of the exploration cost (time) in SANE, we include the human assessment during the initialization only where the quality of the initial samples can be accessed (good or bad) via visualization from domain experts. To transform the assessment into a quantitative metric, we compute the mean distance $d_f$ between selected locations and all feasible assessed locations $X_{i\_good}$, and mean distance $d_{if}$ between a selected location and all infeasible assessed locations $X_{i\_bad}$ respectively. It is to be noted $X_{i\_good}$ and $X_{i\_bad}$ combines the initial sampled locations $D_{X_0}$. Then we compute the constraint function as per eq. 10, where the positive value indicates feasibility and negative value indicates infeasibility.

$$c \geq 0 => d_{if} - d_f \geq 0 \tag{10}$$

$$d_f = \frac{\sum_{i\_good=1}^{n\_good} |X - X_{i\_good}|}{n\_good}, \; X_{i\_good} \subseteq D_{X_0} \tag{11}$$

$$d_{if} = \frac{\sum_{i\_bad=1}^{n\_bad} |X - X_{i\_bad}|}{n\_bad}, X_{i\_bad} \subseteq D_{X_0} \tag{12}$$

$$D_{X_0} = \{X_{i_{good}}, X_{i_{bad}}\} \tag{13}$$

After the initialization as the iteration $i$ progresses, the new samples are not assessed as this would increase the cost of the SANE workflow. In other words, it is not an appropriate approach for the experimentalist to assess the quality of every SANE navigated sample, as this will completely diminish the purpose of an autonomous workflow. The goal for any human or domain intervened autonomous system would be at the level of minimal intervention with maximum improvement. Thus, the human intervention in the gated-SANE is limited to initialization. This initialized assessed gate constraint data $c$ is fitted to a surrogate model $\triangle_c$, and after training, the mean estimation $\bar{c}$ is calculated for all the locations in the parameter space to define the mean estimated gate constraint map. With newly explored locations, the training data is augmented, the gate-surrogate model $\triangle_c$ is retrained, and the mean estimated gate constraint map is updated. Then, the gate constraint is linked to the strategic acquisition function with a penalty factor $P$ as per eq. 14. $\beta$ is the order of magnitude (ceiling factor) of the maximum value of $u_c(X)$, to avoid the imbalance order of magnitude between $u_c(X)$ and $\bar{c}(X)$.

$$u_c(X) = \begin{cases} u_c(X), & \bar{c}(X) \geq 0 \\ u_c(X) + P \times \beta \times \bar{c}(X), & \bar{c}(X) < 0 \end{cases} \tag{14}$$

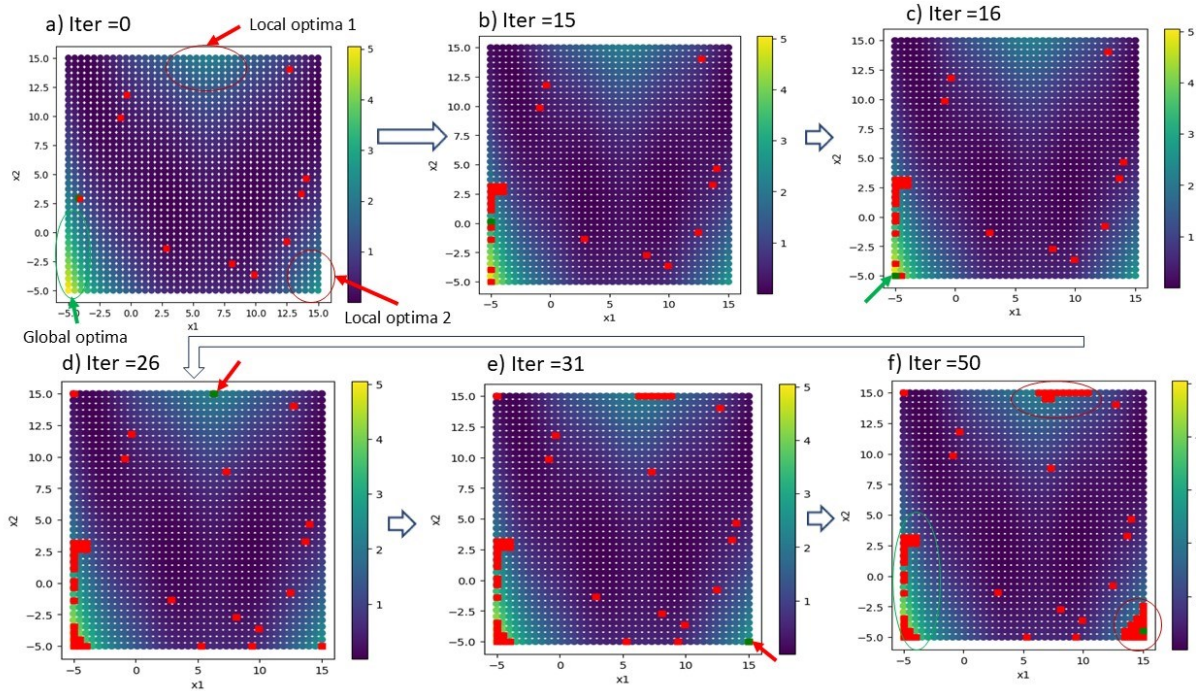

**Figure 2.** Overview of navigation of Strategic Autonomous Non-Smooth Exploration (SANE) on a synthetic 2D search space. Here, the ground truth has one global optimal and 2 local minima. 2a) is the state of initialization with 10 starting samples randomly selected (denoted by red dots). 2b) After 15 iterations, we find the global optimal region and therefore 2c) shows the discovery of the global optimal point (denoted by green dot). 2d) shows the exploitation of the global region of interest and discovery of the first local region of interest (denoted by green dot) following the probabilistic approach as per Eqs. 1- 7. 2e) shows the exploitation of the first local region of interest and discovery of the second local region of interest (denoted by green dot). 2f) shows the final strategic explored map at iteration 50 with exploitation of both the global and local region of interests.

To validate the SANE workflow, we attempt to implement over a synthetic 2D search space. The 2D test function is the multimodal Branin function[51], containing one global optimal and two local optima. The formulation of the function is provided in the Supplementary Material. The prediction model for data $\Delta_D$ is considered as Gaussian process (GP) regression model with Radial Basis kernel function (RBF), and the standard acquisition function $u(X)$ is considered as the Expected Improvement (EI). The detailed formulation of the GP model and EI acquisition function is provided in the Supplementary Material. Here, the switching parameter trajectory is set as $s_i = 0, i < 25$ and $s_i = 1, 25 < i \leq 50$. Here the total number of iterations $N = 50$. Throughout the paper we consider checking for a new region of interest after every $n = 5$ iterations. It is evident to say that, since this is non-physical synthetic data, no human-intervened constraint is relevant and therefore we ignore the gate implementation in this example. **Fig 2**. shows the overview of the strategic navigation from SANE, to discover and exploit the global and local region of interests. **Fig 3**. shows the comparison of the exploration between proposed SANE and standard BO, with two initializations. The initial samples for Run 1 are similar to as in Fig 2a) where the initial samples are far away from all the region of interest. Compared with BO (fig.3a), we can clearly see SANE (fig.3b) provides better exploration with high volume of exploitation in all the regions of interest, and better-balanced exploitation among those regions of interest. Contrastingly, BO only focused on exploiting the global region of interest and explores many uninteresting locations rather than exploiting local regions. We see estimated function map has a good agreement with the ground truth as well, which again signifies the redundant sampling of BO and better navigation strategy of SANE. In other words, unlike in standard BO, SANE spends more time on finding insights in good region, while it still performs a good agreement with the overall ground truth. We see similar interpretations with the second initialization, where at least one of the initial samples is within each of the regions of interest (refer to **Fig.S1.** in the Supplementary Materials). Here also the standard BO did not exploit in the local region (fig.3c) whereas the SANE did (fig. 3d).

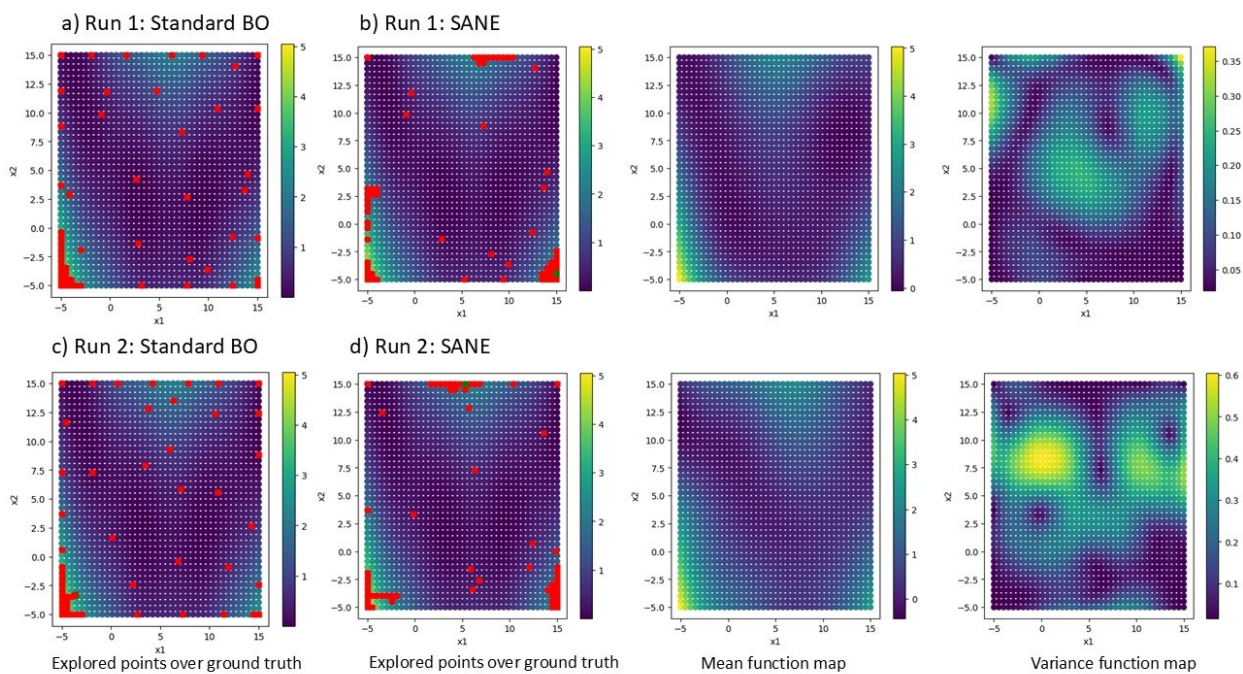

**Figure 3.** Comparison of exploration between BO and SANE on a synthetic 2D search space at two different random initializations. In figures (b) and (d), the left figures are the explored samples over ground truth, the middle figures are the GP predicted mean and the right figures are the GP uncertainty.

## 2. Implementing in piezoresponse spectroscopy of composition spread combinatorial library Sm-BFO

We first implemented the strategic BO in a combinatorial library ferroelectric Sm-doped BiFeO3 (Sm-BFO) data[52], details regarding materials of this dataset have been reported in our previous work[44]. Briefly, the dataset, acquired by our AEcroscopy platform[53] with a NanoSurf Driven AFM, comprises a spectrum of piezoresponse vs. voltage hysteresis loops across the composition spread of Sm-BFO. The Sm-BFO comprises transition with an increase of Sm content from the ferroelectric state BiFeO3 to a non-ferroelectric state of 20% Sm-doped BiFeO3; as such, the hysteresis loops corresponding 15%-20% Sm-doped BFO are mostly closed, without much information regarding ferroelectric characteristics. The physical scalarizers extracted from these closed hysteresis loops consist of very high noise, potentially misleading the autonomous discovery. In this work, we consider two such scalarizers: 1) nucleation voltage and 2) coercive voltage, where we aim to minimize those in the exploration strategy. In other words, we aim to locate regions which has lower values of nucleation and coercive voltage. These parameters are critical for ferroelectric materials as they determine the electric field required to switch the polarization direction and initiate domain formation, which influence the materials' energy efficiency, switching speed, and overall applicability in e.g., memory devices and sensors.

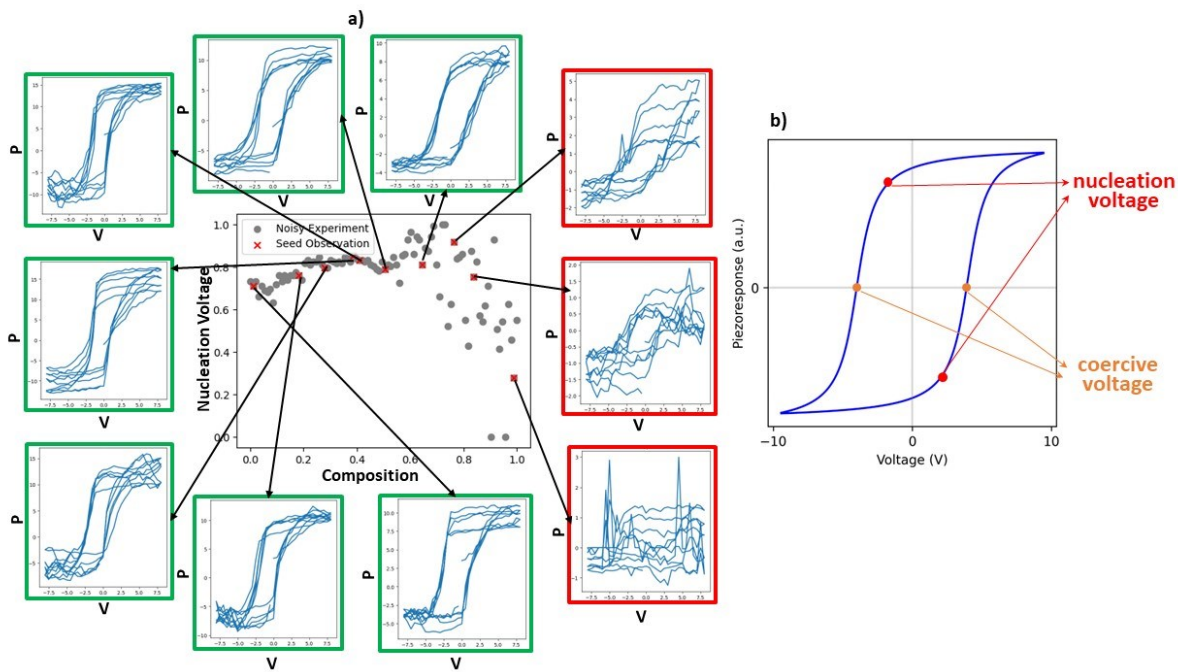

**Figure 4.** Human assessment of the initial samples to define the feasibility of the noisy experiments. In fig. (a), the spectral structure images highlighted in green are the positive assessments while the spectral structure images highlighted in red are the negative assessments. These initial samples are generated via Latin Hypercube sampling method. Fig (b) defines the scalarizes such as nucleation and coercive voltages, given the spectral structure of hysteresis loop.

During the initialization of SANE, where the initial samples are chosen from Latin Hypercube sampling, we formulate the human knowledge driven constrained gate (refer to eqs. 10-14) through voting as shown in **Fig. 4a.** To attain meaningful scalarizer values of nucleation and coercive voltages, an appropriate spectral structure is necessary (**fig. 4b**). Based on this domain knowledge, the assessment of the initial samples has been conducted as per fig. 4a. We can clearly see the role of human intervention in passing knowledge to the ML policy regarding the feasibility of the experimental results, as the spectral of the best (over lowest nucleation voltage) training sample does not form a hysteresis loop to consider a feasible structure. To obtain broad coverage of the parameter space during initialization, we utilized the Latin hypercube sampling technique instead of random sampling. Here, the prediction models for data $\Delta_D$ and gate $\Delta_C$ are considered as Gaussian process (GP) regression models with a Matern kernel function (refer to Supplementary Material), and the constraint penalty factor is chosen as $P = 1000$ through exhaustive analysis with different penalty factors. The standard acquisition function $u(X)$ is considered as the Expected Improvement (EI). Here, the switching parameter trajectory is set as $s_i = 0, i < 15$ and $s_i = 1, 15 < i \leq 30$, where the total number of iterations, $N = 30$. **Fig. 5.** shows the comparison of the autonomous exploration between standard BO and SANE. We can clearly see (figs. b, e) that the pure data-driven BO suggest exploring significantly over the noisy region in the parameter space due to inaccurately measured low values of nucleation and coercive voltages, with insufficient exploitation of the true global region and non-exploitation of local regions. Here in SANE, the dynamic gate (figs. c, f) avoids exploration of those human defined infeasible regions and exploits the feasible global region (highlighted by the green dashed circle). Secondly, we can also see the exploitation of the local regions (pointed with black arrows) to aim to provide deeper insight into the material behavior. One of which would be the understanding of the local robustness of the material behavior, given the physically relevance experimental measurements (feasible spectral structure). For Case 1, in fig. (c), we can see comparing two local regions of interest, the left local region (pointed by solid black arrow) gives lower nucleation voltage and lower local fluctuation/noise than the right local region (pointed by dashed black arrow). Case 2 observation has more interesting trade-offs where, in fig. (f), we can see comparing two local regions of interest, the left local region (pointed by solid black arrow) gives higher coercive voltage but lower local fluctuation/noise than the right local region (pointed by dashed black arrow). Also, the right local region is very close to the estimated infeasible space. We can see from the standard BO, this local exploitation has not been suggested within the same cost of exploration, due to being overly exploring on infeasible region and ignorance of potentially interesting local regions.

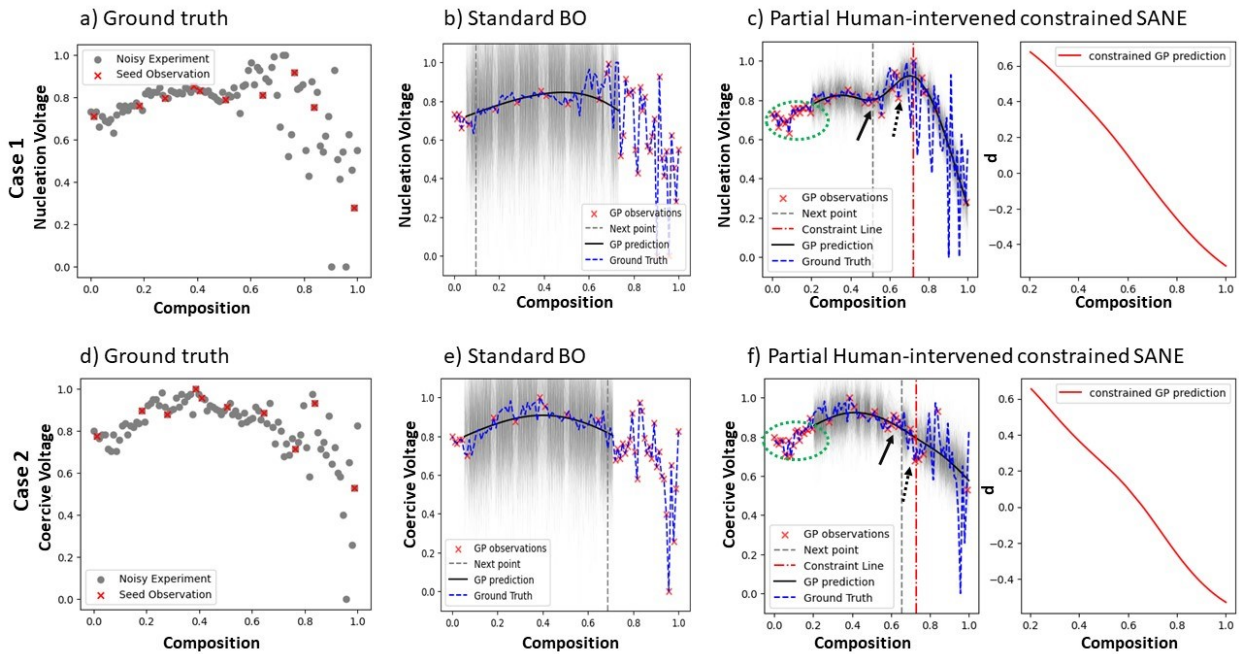

**Figure 5.** Comparison of standard BO and SANE exploration in a combinatorial library ferroelectric Sm-doped BiFeO3 (Sm-BFO) data. In Case 1, the objective is the minimization of the nucleation voltage and in Case 2, the objective is the minimization of the Coercive voltage. Figs (a) and (d) are the noisy experimental ground truth data for these two cases, where the red dots are the initially LHS-driven selected samples. Figs (b) and (c) are the standard BO and the SANE exploration respectively for Case 1, and figs (e) and (f) are the same plots for Case 2, where the black solid lines are the respective estimated mean of the ground truth function, and the black shaded regions are the uncertainty. The red "x" points are the 30 autonomously driven samples for each case study. The red vertical dashed line in the left figs of (c) and (f) are the estimated constraint boundaries or gate $\bar{c} = 0$ where the left region of the boundary is the feasible region $\bar{c} > 0$, and the right region of the boundary is the infeasible region $\bar{c} < 0$. The right figs of (c) and (f) are the estimated constrained gate function $\bar{c}$ with feasible region of $\bar{c} > 0$.

### 3. Implementing in BEPS PTO data

The second model dataset is a vertical band excitation piezoresponse spectroscopy (BEPS) data of a PbTiO3 (PTO) thin film.[45] Here, we aim to locate regions which have higher values of loop area for advanced memory device application. We initialize the SANE with 30 LHS driven initial samples and their domain expert assessments. Here, the prediction models for data $\Delta_D$ and gate $\Delta_C$ are considered as deep kernel learning (DKL) regression models[54] with Matern kernel and RBF kernel functions (refer to Supplementary Material), and the constraint penalty factor is chosen as $P = 1000$. For high dimensionality problems, deep kernel provides better estimation of the structure-property relationship than a standard GP.[55] DKL is built on the framework on fully-connected neural network (NN) where the high-dimensional input image patch is first embedded into low dimensional kernel space (in this case set as 2), and then a standard GP kernel operates, such that the parameters of GP and weights of NN are learned jointly. This DKL technique has been implemented for better exploration through active learning in experimental environments.[56–60] Here, we utilized a DKL implementation from open-source Gpax software package. Here, for a selected sample co-ordinate as suggested by the acquisition function, we input a local structure image patch (high dimensional data) to the DKL which predicts the output (loop area). Unlike in the standard GP, through DKL we can provide prior knowledge about the local correlation of the structural image patches. The standard acquisition function $u(X)$ is considered as the Expected Improvement (EI). Here, the switching parameter trajectory is set as $s_i = 0, i < 60$ and $s_i = 1, 60 < i \leq 100$, where the total number of iterations, $N = 100$.

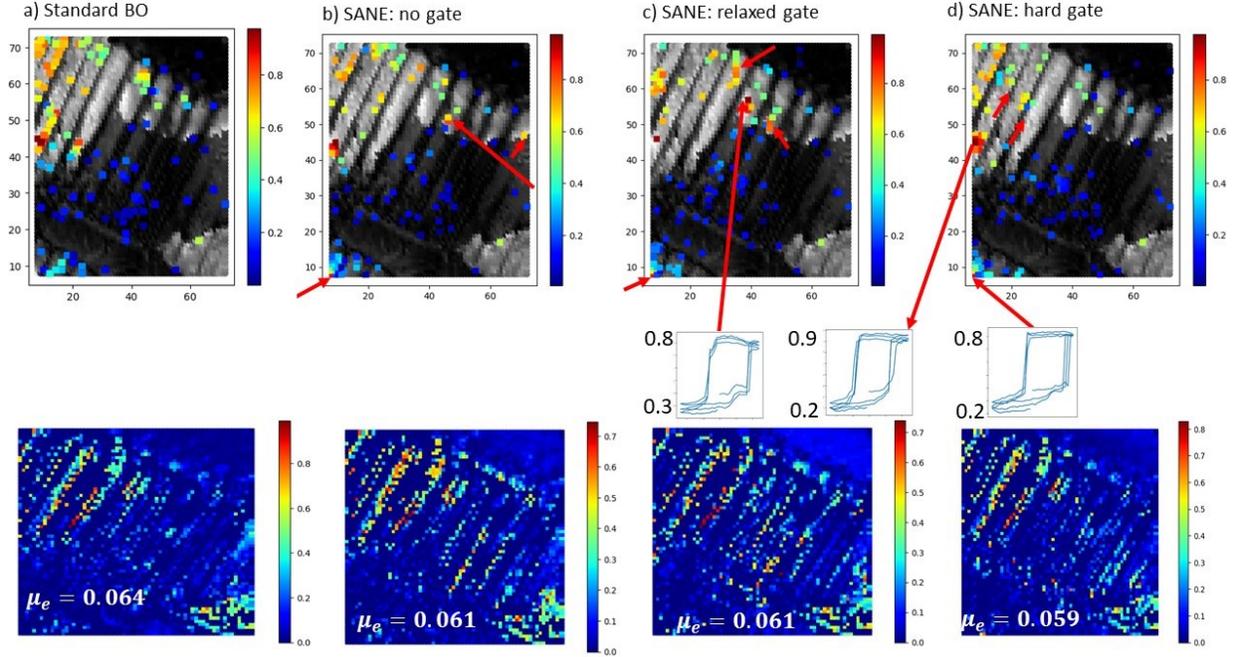

**Figure 6.** Comparison of standard BO and SANE exploration in BEPS data of a PbTiO3 (PTO) thin film Fig (a) is the exploration from standard BO, fig(b) is the SANE exploration without implementation of the human-intervened gate, fig(c) is the SANE exploration with relaxed human-intervened gate, fig(d) is the SANE exploration with hard human-intervened gate. For each subfigure, the top figure is the explored points over ground truth (loop area) after 100 iterations. The color of the explored points indicates the loop area value where red indicates the higher values (optimal regions) and blue indicates the lower values. The bottom figure is the absolute error map over only the feasible region between the ground truth and the prediction with mean values of $\mu_e = 0.064$, $\mu_e = 0.061$, of $\mu_e = 0.061$, of $\mu_e = 0.059$ respectively.

**Fig. 6.** compares the performance between the standard BO exploration and SANE exploration after 100 iterations. On the SANE exploration, we considered three workflows- 1) without gate, 2) with a relaxed gate and 3) with the hard gate. The relaxed gate is designed with only estimation of the constrained gate map with initial data and is not being updated with selection of new samples. The hard gate is designed as described earlier. The constrained gate maps are provided in the Supplementary **Fig. S2**. Comparing with BO, we can clearly see the diversification of the search in all SANE exploration where the standard BO concentrated mostly on a single region (top left boundary of the image space), assuming no potential good solutions can be found in other areas. Whereas all the SANE workflows discover another optimal region (left bottom corner of the image space indicated with red arrow) where we can see the desired spectral structure with large loop area. Among all the SANE workflows, the relaxed constrained SANE achieves discovery of another optimal region where we can see the desired spectra (indicated by red arrow in fig. 6c) with a large loop area. With further investigation, we understand that the region is very close to infeasibility for the estimated hard gate and is therefore not being explored. SANE explores a lot of infeasible regions over the unconstrained search space and therefore failed to locate the true feasible optimal region. This shows the role of human intervention to reduce the search space and subsequently, intelligent co-navigation to locate the optimal regions in the narrow space. However, we need to ensure the proper balancing of the location of the gates and therefore proper tuning of penalty factor *P* is required, which is problem dependent. Comparing the mean absolute error between the ground truth and the prediction over the estimated feasible space, defined by the hard gate, we can see SANE has lower error than BO. However, it is to be noted that SANE is not designed to guarantee to always minimize the prediction of the overall feasible multi-modal or non-differentiable parameter space than BO, which is a task for the prediction model. Here, the purpose of SANE is to reduce the over dependencies of the prediction model and explore strategically to avoid being stuck in a single region and probabilistically hop to explore other region of interest, when the parameter space is too complex due to noisy experiments. To design a better predictive model to learn a complex non-differential function is a future scope.

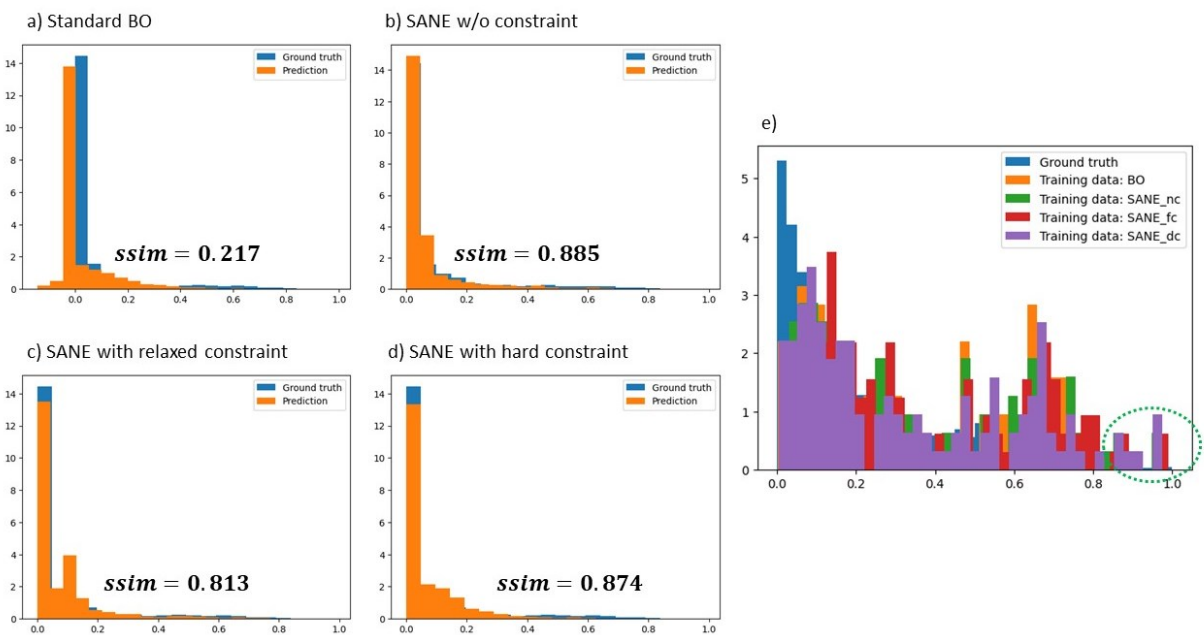

**Figure 7.** Comparison of standard BO and SANE exploration in BEPS data of a PbTiO3 (PTO) thin film. Figs (a)-(d) are the comparison of the histogram between the ground truth and the prediction at the estimated feasible region; driven by standard BO, SANE without implementation of the human-intervened gate, SANE with relaxed human-intervened gate, and SANE with hard human-intervened gate. Fig (e) is the comparison chart of the histograms among ground truth samples (blue), BO-driven samples (orange), SANE w/o constrained gate driven samples (green), SANE with relaxed constrained gate driven samples (red) and SANE with hard constrained gate driven samples (purple).

For further performance validation of SANE, we compare the histograms between the ground truth and the predictions over the estimated feasible space and compare the ground truth with the explored samples after 100 iterations as in **Fig. 7.** From the histogram of the ground truth, we can clearly see the majority of the areas are non-interesting regions with narrow patches of scattered local and global optimal regions. Along with the noisy experiment (fake optimal), the low ratio of the true region of interest makes the exploration even harder, like finding needle in a haystack. That is understandable on why standard BO focus on one region of interest and could not find suitable learning to exploit other regions. Therefore, we can see it fails to provide a better overall prediction at the estimated feasible region where it has the lowest ($ssim_{BO} = 0.217$) similarity to the ground truth histogram (ref to fig 7a). The similarity index for all the SANE workflows (figs. 7b, c, d) are much higher and comparable with each other. From fig. 7e, we can see all the SANE workflows overall explore a better ratio of different optimal regions, as it locates higher percentage of spectral having higher loop areas (indicated by green circle). This is supported with the analysis in fig. 6 where SANE discovers multiple optimal regions. We can clearly see BO fails to do so as it mostly explores near the single region of interest and ending up locating spectral with loop areas ranging between 0.5 to 0.7.

**Conclusion**

In conclusion, we develop a Strategic Autonomous Non-Smooth Exploration (SANE) framework, which demonstrates advancements in exploring multidimensional parameter spaces with multi-modal and non-differentiable black box functions for materials and physics discovery. Traditional BO methods, while powerful, often lead to the risk of over-focusing on singular optimal conditions and the potential for becoming trapped in noisy regions or local optima. SANE integrates a cost-driven probabilistic acquisition function, a more robust and exploratory approach, to address these limitations in autonomous experimentation. SANE actively seeks out multiple global and local optima, ensuring a more comprehensive exploration of the parameter space. The application of the SANE framework in two complex material systems, *i.e.,* Sm-doped $BiFeO_3$ combinatorial library and $PbTiO_3$ ferroelastic/ferroelectric thin films, has demonstrated its efficacy. In both cases, SANE outperformed traditional BO by avoiding entrapment in noisy regions and/or singular optimum, enabling the discovery of multiple optimal conditions and uncovering a broader spectrum of material behaviors. Moreover, by incorporating a dynamic constrained gate driven by human domain knowledge, we have further enhanced the SANE framework by prioritizing scientifically valuable regions of the parameter space. This approach not only mitigates the challenges posed by experimental noise and uncertainties but also ensures the exploration aligns with broader scientific discovery goals.

As autonomous experimentation for accelerated research continues to evolve, the integration of advanced autonomous exploration methods like SANE, coupled with human expertise, will stand as a critical tool for pushing the boundaries of what is possible in autonomous research, more comprehensive explorations of complex material landscapes and driving new scientific discoveries. The developed approach here can be applied in a broad materials science field including materials synthesis, characterization, and computation, offering a more comprehensive and effective path forward in autonomous research.

**Additional Information:**

See the supplementary material for additional analysis and figures, related to the research.

**Acknowledgements:**

This work (A.B) was supported by the University of Tennessee startup funding. The authors acknowledge the use of facilities and instrumentation at the UT Knoxville Institute for Advanced Materials and Manufacturing (IAMM) and the Shull Wollan Center (SWC) supported in part by the National Science Foundation Materials Research Science and Engineering Center program through the UT Knoxville Center for Advanced Materials and Manufacturing (DMR-2309083). This effort (PFM datasets) is supported by the Center for Nanophase Materials Sciences (CNMS), which is a US Department of Energy, Office of Science User Facility at Oak Ridge National Laboratory. This work (R.V. and I.T.) were supported by DOE EFRC 3DFEM[2]. This work (H.F) was supported by Japan Science and Technology Agency (JST) as part of Adopting Sustainable Partnerships for Innovative Research Ecosystem (ASPIRE), Grant Number JPMJAP2312.

**Conflict of Interest:**

The author confirms no conflict of interest.

**Authors Contribution**

A.B. and Y.L. conceived the project. A.B. developed SANE framework. Y.L and R.V developed AEcroscopy platform for data acquisition. Y.L. acquired experimental data for SANE investigation and A.B. performed the analysis. R.P and I.T prepared Sm-doped BiFeO3 sample. H.F prepared PbTiO3 (PTO) thin film sample. A.B. and Y.L. wrote the manuscript, all authors edited the manuscript.

**Code and Data Availability Statement:**

The analysis reported here along with the code is summarized in Colab Notebook for the purpose of tutorial and application to other data and can be found in https://github.com/arpanbiswas52/SANE

**References:**

(1) Abolhasani, M.; Kumacheva, E. The Rise of Self-Driving Labs in Chemical and Materials Sciences. *Nat. Synth.* **2023**, *2* (6), 483–492. https://doi.org/10.1038/s44160-022-00231-0.
(2) Kalinin, S. V.; Ziatdinov, M.; Hinkle, J.; Jesse, S.; Ghosh, A.; Kelley, K. P.; Lupini, A. R.; Sumpter, B. G.; Vasudevan, R. K. Automated and Autonomous Experiments in Electron and Scanning Probe Microscopy. *ACS Nano* **2021**, *15* (8), 12604–12627. https://doi.org/10.1021/acsnano.1c02104.
(3) Stach, E.; DeCost, B.; Kusne, A. G.; Hattrick-Simpers, J.; Brown, K. A.; Reyes, K. G.; Schrier, J.; Billinge, S.; Buonassisi, T.; Foster, I.; Gomes, C. P.; Gregoire, J. M.; Mehta, A.; Montoya, J.; Olivetti, E.; Park, C.; Rotenberg, E.; Saikin, S. K.; Smullin, S.; Stanev, V.; Maruyama, B. Autonomous Experimentation Systems for Materials Development: A Community Perspective. *Matter* **2021**, *4* (9), 2702–2726. https://doi.org/10.1016/j.matt.2021.06.036.
(4) Tom, G.; Schmid, S. P.; Baird, S. G.; Cao, Y.; Darvish, K.; Hao, H.; Lo, S.; Pablo-García, S.; Rajaonson, E. M.; Skreta, M.; Yoshikawa, N.; Corapi, S.; Akkoc, G. D.; Strieth-Kalthoff, F.; Seifrid, M.; Aspuru-Guzik,

A. Self-Driving Laboratories for Chemistry and Materials Science. ChemRxiv June 18, 2024. https://doi.org/10.26434/chemrxiv-2024-rj946-v2.

(5) Xie, Y.; Sattari, K.; Zhang, C.; Lin, J. Toward Autonomous Laboratories: Convergence of Artificial Intelligence and Experimental Automation. *Prog. Mater. Sci.* **2023**, *132*, 101043. https://doi.org/10.1016/j.pmatsci.2022.101043.

(6) Shahriari, B.; Swersky, K.; Wang, Z.; Adams, R. P.; de Freitas, N. Taking the Human Out of the Loop: A Review of Bayesian Optimization. *Proc. IEEE* **2016**, *104* (1), 148–175. https://doi.org/10.1109/JPROC.2015.2494218.

(7) Jones, D. R.; Schonlau, M.; Welch, W. J. Efficient Global Optimization of Expensive Black-Box Functions. *J. Glob. Optim.* **1998**, *13* (4), 455–492. https://doi.org/10.1023/A:1008306431147.

(8) Biswas, A.; Hoyle, C. An Approach to Bayesian Optimization for Design Feasibility Check on Discontinuous Black-Box Functions. *J. Mech. Des.* **2021**, *143* (031716). https://doi.org/10.1115/1.4049742.

(9) Quadrianto, N.; Kersting, K.; Xu, Z. Gaussian Process. In *Encyclopedia of Machine Learning*; Sammut, C., Webb, G. I., Eds.; Springer US: Boston, MA, 2010; pp 428–439. https://doi.org/10.1007/978-0-387-30164-8_324.

(10) Deringer, V. L.; Bartók, A. P.; Bernstein, N.; Wilkins, D. M.; Ceriotti, M.; Csányi, G. Gaussian Process Regression for Materials and Molecules. *Chem. Rev.* **2021**, *121* (16), 10073–10141. https://doi.org/10.1021/acs.chemrev.1c00022.

(11) Noack, M. M.; Doerk, G. S.; Li, R.; Streit, J. K.; Vaia, R. A.; Yager, K. G.; Fukuto, M. Autonomous Materials Discovery Driven by Gaussian Process Regression with Inhomogeneous Measurement Noise and Anisotropic Kernels. *Sci. Rep.* **2020**, *10* (1), 17663. https://doi.org/10.1038/s41598-020-74394-1.

(12) Brochu, E.; Cora, V. M.; de Freitas, N. A Tutorial on Bayesian Optimization of Expensive Cost Functions, with Application to Active User Modeling and Hierarchical Reinforcement Learning. arXiv December 12, 2010. https://doi.org/10.48550/arXiv.1012.2599.

(13) Cox, D. D.; John, S. A Statistical Method for Global Optimization. In *[Proceedings] 1992 IEEE International Conference on Systems, Man, and Cybernetics*; 1992; pp 1241–1246 vol.2. https://doi.org/10.1109/ICSMC.1992.271617.

(14) Jones, D. R. A Taxonomy of Global Optimization Methods Based on Response Surfaces. *J. Glob. Optim.* **2001**, *21* (4), 345–383. https://doi.org/10.1023/A:1012771025575.

(15) Kushner, H. J. A New Method of Locating the Maximum Point of an Arbitrary Multipeak Curve in the Presence of Noise. *J. Basic Eng.* **1964**, *86* (1), 97–106. https://doi.org/10.1115/1.3653121.

(16) Sanchez, S. L.; Foadian, E.; Ziatdinov, M.; Yang, J.; Kalinin, S. V.; Liu, Y.; Ahmadi, M. Physics-Driven Discovery and Bandgap Engineering of Hybrid Perovskites. *Digit. Discov.* **2024**, *3* (8), 1577–1590. https://doi.org/10.1039/D4DD00080C.

(17) Chang, J.; Nikolaev, P.; Carpena-Núñez, J.; Rao, R.; Decker, K.; Islam, A. E.; Kim, J.; Pitt, M. A.; Myung, J. I.; Maruyama, B. Efficient Closed-Loop Maximization of Carbon Nanotube Growth Rate Using Bayesian Optimization. *Sci. Rep.* **2020**, *10* (1), 9040. https://doi.org/10.1038/s41598-020-64397-3.

(18) Ueno, T.; Rhone, T. D.; Hou, Z.; Mizoguchi, T.; Tsuda, K. COMBO: An Efficient Bayesian Optimization Library for Materials Science. *Mater. Discov.* **2016**, *4*, 18–21. https://doi.org/10.1016/j.md.2016.04.001.

(19) Kalinin, S. V.; Ziatdinov, M.; Vasudevan, R. K. Guided Search for Desired Functional Responses via Bayesian Optimization of Generative Model: Hysteresis Loop Shape Engineering in Ferroelectrics. *J. Appl. Phys.* **2020**, *128* (2), 024102. https://doi.org/10.1063/5.0011917.

(20) Biswas, A.; Morozovska, A. N.; Ziatdinov, M.; Eliseev, E. A.; Kalinin, S. V. Multi-Objective Bayesian Optimization of Ferroelectric Materials with Interfacial Control for Memory and Energy Storage Applications. *J. Appl. Phys.* **2021**, *130* (20), 204102. https://doi.org/10.1063/5.0068903.

(21) Morozovska, A. N.; Eliseev, E. A.; Biswas, A.; Shevliakova, H. V.; Morozovsky, N. V.; Kalinin, S. V. Chemical Control of Polarization in Thin Strained Films of a Multiaxial Ferroelectric: Phase Diagrams and Polarization Rotation. *Phys. Rev. B* **2022**, *105* (9), 094112. https://doi.org/10.1103/PhysRevB.105.094112.

(22) Morozovska, A. N.; Eliseev, E. A.; Biswas, A.; Morozovsky, N. V.; Kalinin, S. V. Effect of Surface Ionic Screening on Polarization Reversal and Phase Diagrams in Thin Antiferroelectric Films for Information and Energy Storage. *Phys. Rev. Appl.* **2021**, *16* (4), 044053. https://doi.org/10.1103/PhysRevApplied.16.044053.

(23) Tao, S.; van Beek, A.; Apley, D. W.; Chen, W. Multi-Model Bayesian Optimization for Simulation-Based Design. *J. Mech. Des.* **2021**, *143* (111701). https://doi.org/10.1115/1.4050738.

(24) Narasimha, G.; Hus, S.; Biswas, A.; Vasudevan, R.; Ziatdinov, M. Autonomous Convergence of STM Control Parameters Using Bayesian Optimization. *APL Mach. Learn.* **2024**, *2* (1), 016121. https://doi.org/10.1063/5.0185362.

(25) Liu, Y.; Kelley, K. P.; Vasudevan, R. K.; Funakubo, H.; Ziatdinov, M. A.; Kalinin, S. V. Experimental Discovery of Structure–Property Relationships in Ferroelectric Materials via Active Learning. *Nat. Mach. Intell.* **2022**, *4* (4), 341–350. https://doi.org/10.1038/s42256-022-00460-0.

(26) Griffiths, R.-R.; Hernández-Lobato, J. M. Constrained Bayesian Optimization for Automatic Chemical Design Using Variational Autoencoders. *Chem. Sci.* **2020**, *11* (2), 577–586. https://doi.org/10.1039/C9SC04026A.

(27) Burger, B.; Maffettone, P. M.; Gusev, V. V.; Aitchison, C. M.; Bai, Y.; Wang, X.; Li, X.; Alston, B. M.; Li, B.; Clowes, R.; Rankin, N.; Harris, B.; Sprick, R. S.; Cooper, A. I. A Mobile Robotic Chemist. *Nature* **2020**, *583* (7815), 237–241. https://doi.org/10.1038/s41586-020-2442-2.

(28) Kusne, A. G.; Yu, H.; Wu, C.; Zhang, H.; Hattrick-Simpers, J.; DeCost, B.; Sarker, S.; Oses, C.; Toher, C.; Curtarolo, S.; Davydov, A. V.; Agarwal, R.; Bendersky, L. A.; Li, M.; Mehta, A.; Takeuchi, I. On-the-Fly Closed-Loop Materials Discovery via Bayesian Active Learning. *Nat. Commun.* **2020**, *11* (1), 5966. https://doi.org/10.1038/s41467-020-19597-w.

(29) Dave, A.; Mitchell, J.; Burke, S.; Lin, H.; Whitacre, J.; Viswanathan, V. Autonomous Optimization of Non-Aqueous Li-Ion Battery Electrolytes via Robotic Experimentation and Machine Learning Coupling. *Nat. Commun.* **2022**, *13* (1), 5454. https://doi.org/10.1038/s41467-022-32938-1.

(30) Harris, S. B.; Biswas, A.; Yun, S. J.; Roccapriore, K. M.; Rouleau, C. M.; Puretzky, A. A.; Vasudevan, R. K.; Geohegan, D. B.; Xiao, K. Autonomous Synthesis of Thin Film Materials with Pulsed Laser Deposition Enabled by In Situ Spectroscopy and Automation. *Small Methods n/a* (n/a), 2301763. https://doi.org/10.1002/smtd.202301763.

(31) Kanarik, K. J.; Osowiecki, W. T.; Lu, Y. (Joe); Talukder, D.; Roschewsky, N.; Park, S. N.; Kamon, M.; Fried, D. M.; Gottscho, R. A. Human–Machine Collaboration for Improving Semiconductor Process Development. *Nature* **2023**, *616* (7958), 707–711. https://doi.org/10.1038/s41586-023-05773-7.

(32) Biswas, A.; Liu, Y.; Creange, N.; Liu, Y.-C.; Jesse, S.; Yang, J.-C.; Kalinin, S. V.; Ziatdinov, M. A.; Vasudevan, R. K. A Dynamic Bayesian Optimized Active Recommender System for Curiosity-Driven Partially Human-in-the-Loop Automated Experiments. *Npj Comput. Mater.* **2024**, *10* (1), 1–12. https://doi.org/10.1038/s41524-023-01191-5.

(33) Biswas, A.; Liu, Y.; Ziatdinov, M.; Liu, Y.-C.; Jesse, S.; Yang, J.-C.; Kalinin, S.; Vasudevan, R. A Multi-Objective Bayesian Optimized Human Assessed Multi-Target Generated Spectral Recommender System for Rapid Pareto Discoveries of Material Properties; American Society of Mechanical Engineers Digital Collection, 2023. https://doi.org/10.1115/DETC2023-116956.

(34) Liu, Y.; Ziatdinov, M. A.; Vasudevan, R. K.; Kalinin, S. V. Explainability and Human Intervention in Autonomous Scanning Probe Microscopy. *Patterns* **2023**, *4* (11), 100858. https://doi.org/10.1016/j.patter.2023.100858.

(35) Kalinin, S. V.; Liu, Y.; Biswas, A.; Duscher, G.; Pratiush, U.; Roccapriore, K.; Ziatdinov, M.; Vasudevan, R. Human-in-the-Loop: The Future of Machine Learning in Automated Electron Microscopy. *Microsc. Today* **2024**, *32* (1), 35–41. https://doi.org/10.1093/mictod/qaad096.

(36) Harris, S. B.; Vasudevan, R.; Liu, Y. Active Oversight and Quality Control in Standard Bayesian Optimization for Autonomous Experiments. arXiv May 25, 2024. https://doi.org/10.48550/arXiv.2405.16230.

(37) Biswas, A.; Valleti, S. M. P.; Vasudevan, R.; Ziatdinov, M.; Kalinin, S. V. Towards Accelerating Physical Discovery via Non-Interactive and Interactive Multi-Fidelity Bayesian Optimization: Current Challenges and Future Opportunities. arXiv February 20, 2024. https://doi.org/10.48550/arXiv.2402.13402.

(38) Garnett, R. *Bayesian Optimization*. Cambridge Core. https://doi.org/10.1017/9781108348973.

(39) Balandat, M.; Karrer, B.; Jiang, D. R.; Daulton, S.; Letham, B.; Wilson, A. G.; Bakshy, E. BOTORCH: A Framework for Efficient Monte-Carlo Bayesian Optimization. In *Proceedings of the 34th International Conference on Neural Information Processing Systems*; NIPS '20; Curran Associates Inc.: Red Hook, NY, USA, 2020; pp 21524–21538.

(40) Ziatdinov, M. A.; Liu, Y.; Morozovska, A. N.; Eliseev, E. A.; Zhang, X.; Takeuchi, I.; Kalinin, S. V. Hypothesis Learning in Automated Experiment: Application to Combinatorial Materials Libraries. *Adv. Mater.* **2022**, *34* (20), 2201345. https://doi.org/10.1002/adma.202201345.

(41) Gaudioso, M.; Giallombardo, G.; Miglionico, G. Essentials of Numerical Nonsmooth Optimization. *Ann. Oper. Res.* **2022**, *314* (1), 213–253. https://doi.org/10.1007/s10479-021-04498-y.

(42) Luo, H.; Demmel, J. W.; Cho, Y.; Li, X. S.; Liu, Y. Non-Smooth Bayesian Optimization in Tuning Problems. arXiv September 15, 2021. https://doi.org/10.48550/arXiv.2109.07563.

(43) Slautin, B. N.; Pratiush, U.; Ivanov, I. N.; Liu, Y.; Pant, R.; Zhang, X.; Takeuchi, I.; Ziatdinov, M. A.; Kalinin, S. V. Co-Orchestration of Multiple Instruments to Uncover Structure-Property Relationships in Combinatorial Libraries. *Digit. Discov.* **2024**, *3* (8), 1602–1611. https://doi.org/10.1039/D4DD00109E.

(44) Raghavan, A.; Pant, R.; Takeuchi, I.; Eliseev, E. A.; Checa, M.; Morozovska, A. N.; Ziatdinov, M.; Kalinin, S. V.; Liu, Y. Evolution of Ferroelectric Properties in SmxBi1-xFeO3 via Automated Piezoresponse Force Microscopy across Combinatorial Spread Libraries. arXiv May 14, 2024. https://doi.org/10.48550/arXiv.2405.08773.

(45) Pratiush, U.; Funakubo, H.; Vasudevan, R.; Kalinin, S. V.; Liu, Y. Scientific Exploration with Expert Knowledge (SEEK) in Autonomous Scanning Probe Microscopy with Active Learning. arXiv August 4, 2024. https://doi.org/10.48550/arXiv.2408.02071.

(46) Luong, P.; Nguyen, D.; Gupta, S.; Rana, S.; Venkatesh, S. Adaptive Cost-Aware Bayesian Optimization. *Knowl.-Based Syst.* **2021**, *232*, 107481. https://doi.org/10.1016/j.knosys.2021.107481.

(47) Liang, R.; Hu, H.; Han, Y.; Chen, B.; Yuan, Z. CAPBO: A Cost-Aware Parallelized Bayesian Optimization Method for Chemical Reaction Optimization. *AIChE J.* **2024**, *70* (3), e18316. https://doi.org/10.1002/aic.18316.

(48) Lee, E. H.; Perrone, V.; Archambeau, C.; Seeger, M. Cost-Aware Bayesian Optimization. arXiv March 22, 2020. https://doi.org/10.48550/arXiv.2003.10870.

(49) Slautin, B. N.; Liu, Y.; Funakubo, H.; Vasudevan, R. K.; Ziatdinov, M.; Kalinin, S. V. Bayesian Conavigation: Dynamic Designing of the Material Digital Twins via Active Learning. *ACS Nano* **2024**. https://doi.org/10.1021/acsnano.4c05368.

(50) Slautin, B. N.; Liu, Y.; Funakubo, H.; Kalinin, S. V. Unraveling the Impact of Initial Choices and In-Loop Interventions on Learning Dynamics in Autonomous Scanning Probe Microscopy. *J. Appl. Phys.* **2024**, *135* (15), 154901. https://doi.org/10.1063/5.0198316.

(51) *Branin Function*. https://www.sfu.ca/~ssurjano/branin.html (accessed 2024-08-30).

(52) Fujino, S.; Murakami, M.; Anbusathaiah, V.; Lim, S.-H.; Nagarajan, V.; Fennie, C. J.; Wuttig, M.; Salamanca-Riba, L.; Takeuchi, I. Combinatorial Discovery of a Lead-Free Morphotropic Phase Boundary in a Thin-Film Piezoelectric Perovskite. *Appl. Phys. Lett.* **2008**, *92* (20), 202904. https://doi.org/10.1063/1.2931706.

(53) Liu, Y.; Roccapriore, K.; Checa, M.; Valleti, S. M.; Yang, J.-C.; Jesse, S.; Vasudevan, R. K. AEcroscopy: A Software–Hardware Framework Empowering Microscopy Toward Automated and Autonomous Experimentation. *Small Methods n/a* (n/a), 2301740. https://doi.org/10.1002/smtd.202301740.

(54) Wilson, A. G.; Hu, Z.; Salakhutdinov, R.; Xing, E. P. Deep Kernel Learning. arXiv November 6, 2015. https://doi.org/10.48550/arXiv.1511.02222.

(55) Ziatdinov, M.; Ghosh, A.; Wong, C. Y. (Tommy); Kalinin, S. V. AtomAI Framework for Deep Learning Analysis of Image and Spectroscopy Data in Electron and Scanning Probe Microscopy. *Nat. Mach. Intell.* **2022**, *4* (12), 1101–1112. https://doi.org/10.1038/s42256-022-00555-8.

(56) Roccapriore, K. M.; Kalinin, S. V.; Ziatdinov, M. Physics Discovery in Nanoplasmonic Systems via Autonomous Experiments in Scanning Transmission Electron Microscopy. *Adv. Sci.* **2022**, *9* (36), 2203422. https://doi.org/10.1002/advs.202203422.

(57) Ziatdinov, M.; Liu, Y.; Kalinin, S. V. Active Learning in Open Experimental Environments: Selecting the Right Information Channel(s) Based on Predictability in Deep Kernel Learning. arXiv March 18, 2022. https://doi.org/10.48550/arXiv.2203.10181.

(58) Liu, Y.; Kelley, K. P.; Vasudevan, R. K.; Zhu, W.; Hayden, J.; Maria, J.-P.; Funakubo, H.; Ziatdinov, M. A.; Trolier-McKinstry, S.; Kalinin, S. V. Automated Experiments of Local Non-Linear Behavior in Ferroelectric Materials. *Small* **2022**, *18* (48), 2204130. https://doi.org/10.1002/smll.202204130.

(59) Liu, Y.; Kelley, K. P.; Vasudevan, R. K.; Funakubo, H.; Ziatdinov, M. A.; Kalinin, S. V. Experimental Discovery of Structure–Property Relationships in Ferroelectric Materials via Active Learning. *Nat. Mach. Intell.* **2022**, *4* (4), 341–350. https://doi.org/10.1038/s42256-022-00460-0.

(60) Liu, Y.; Vasudevan, R. K.; Kelley, K. P.; Funakubo, H.; Ziatdinov, M.; Kalinin, S. V. Learning the Right Channel in Multimodal Imaging: Automated Experiment in Piezoresponse Force Microscopy. arXiv February 13, 2023. https://doi.org/10.48550/arXiv.2207.03039.

Arpan Biswas[1,a], Rama Vasudevan[2], Rohit Pant[3], Ichiro Takeuchi[3], Hiroshi Funakubo[4], Yongtao Liu[2,b]

[1] University of Tennessee-Oak Ridge Innovation Institute, Knoxville, TN 37996, USA
[2] Center for Nanophase Materials Sciences, Oak Ridge National Laboratory, Oak Ridge, TN 37830, USA
[3] Department of Materials Science and Engineering, University of Maryland, College Park, Maryland 20742, United States of America
[4] Department of Materials Science and Engineering, Tokyo Institute of Technology, Yokohama, 226-8502, Japan

**Appendix A. Additional Formulation**

1. *Branin Function:*

$$f(X) = a(x_2 - bx_1^2 + cx_1 - r^2) + s(1-t)\cos(x_1) + s \tag{S1}$$

Where,

$$a = 1; b = \frac{5.1}{4\pi^2}; c = \frac{5}{\pi}; r = 6; s = 10; t = \frac{1}{8\pi}; -5 \leq x_1, x_2 < 15$$

2. *Gaussian Process (GP) Regression:*

The general form of the GPM is as follows:

$$y(x) = x^T\beta + z(x) \tag{S2}$$

where $x^T\beta$ is the Polynomial Regression model. The polynomial regression model captures the global trend of the data. $z(x)$ is a realization of a correlated Gaussian Process with mean $E[z(x)]$ and covariance $cov(x^i, x^j)$ functions defined as follows:

$$z(x) \sim GP\left(E[z(x)], cov(x^i, x^j)\right); \tag{S3}$$

$$E[z(x)] = 0, cov(x^i, x^j) = \sigma^2 R(x^i, x^j) \tag{S4}$$

$$R(x^i, x^j) = \exp\left(-0.5 * \sum_{m=1}^{d} \frac{(x_m^i - x_m^j)^2}{\theta_m^2}\right); \tag{S5}$$

$$R(x^i, x^j) = \exp\left(-\sqrt{5} \times \sum_{m=1}^{d} \frac{|x_m^i - x_m^j|}{\theta_m}\right) \times \left(1 + \sqrt{5} \times \sum_{m=1}^{d} \frac{|x_m^i - x_m^j|}{\theta_m} + \frac{5}{3} \times \sum_{m=1}^{d} \frac{|x_m^i - x_m^j|^2}{\theta_m^2}\right) \quad (S6)$$

$$\theta_m = (\theta_1, \theta_2, \ldots, \theta_d)$$

where $\sigma^2$ is the overall variance parameter and $\theta_m$ is the correlation length scale parameter in dimension $m$ of $d$ dimension of $x$. These are termed as the hyper-parameters of GP model. $R(x^i, x^j)$ is the spatial correlation function. In this paper, we have considered a Radial Basis function which is given by eqn. S5 and Matern kernel function as per eqn. S6. The objective is to estimate the hyper-parameters $\sigma$, $\theta_m$ which creates the surrogate model that best explains the training data $D_k$ at iteration $k$. In this paper, we used the Monte Carlo Markov Chain (MCMC) approach to estimate the hyperparameters.

After the GP model is fitted, the next task of the GP model is to predict at an arbitrary (unexplored) location drawn from the parameter space. Assume $D_k = \{X_k, Y(X_k)\}$ is the prior information from previous evaluations or experiments from high fidelity models, and $\bar{\bar{x}}_{k+1} \in \bar{\bar{X}}$ is a new design within the unexplored locations in the parameter space, $\bar{\bar{X}}$. The predictive output distribution of $x_{k+1}$, given the posterior GP model, is given by eqn S7.

$$P(\bar{\bar{y}}_{k+1}|D_k, \bar{\bar{x}}_{k+1}, \sigma_k^2, \theta_k) = N(\mu(\bar{\bar{y}}_{k+1}(\bar{\bar{x}}_{k+1})), \sigma^2(\bar{\bar{y}}_{k+1}(\bar{\bar{x}}_{k+1}))) \quad (S7)$$

where:

$$\mu(\bar{\bar{y}}_{k+1}(\bar{\bar{x}}_{k+1})) = cov_{k+1}^T COV_k^{-1} Y_k; \quad (S8)$$

$$\sigma^2(\bar{\bar{y}}_{k+1}(\bar{\bar{x}}_{k+1})) = cov(\bar{\bar{x}}_{k+1}, \bar{\bar{x}}_{k+1}) - cov_{k+1}^T COV_k^{-1} cov_{k+1} \quad (S9)$$

$COV_k$ is the kernel matrix of already sampled designs $X_k$ and $cov_{k+1}$ is the covariance function of new design $\bar{\bar{x}}_{k+1}$ which is defined as follows:

$$COV_k = \begin{bmatrix} cov(x_1, x_1) & \cdots & cov(x_1, x_k) \\ \vdots & \ddots & \vdots \\ cov(x_k, x_1) & \cdots & cov(x_k, x_k) \end{bmatrix}$$

$$cov_{k+1} = [cov(\bar{\bar{x}}_{k+1}, x_1), cov(\bar{\bar{x}}_{k+1}, x_2), \ldots, cov(\bar{\bar{x}}_{k+1}, x_k)]$$

3. *Expected Improvement acquisition function:*

$$EI(\bar{\bar{y}}(\bar{\bar{x}}|f=1)) =$$

$$\begin{cases} (\mu(\bar{\bar{y}}(\bar{\bar{x}})) - y(x^+) - \xi) \times \Phi(Z, 0, 1) + \sigma(\bar{\bar{y}}(\bar{\bar{x}})) \times \phi(Z) & \text{if } \sigma(\bar{\bar{y}}(\bar{\bar{x}})) > 0 \\ 0 \text{ if } \sigma(\bar{\bar{y}}(\bar{\bar{x}})) = 0 \end{cases}$$

(S10)

$$Z = \begin{cases} \frac{\mu(\bar{\bar{y}}(\bar{\bar{x}})) - y(x^+) - \xi}{\sigma(\bar{\bar{y}}(\bar{\bar{x}}))} & \text{if } \sigma(\bar{\bar{y}}(\bar{\bar{x}})) > 0 \\ 0 & \text{if } \sigma(\bar{\bar{y}}(\bar{\bar{x}})) = 0 \end{cases}$$

(S11)

where $y(x^+)$ is the current maximum value among all the sampled data until the current stage; $\mu(\bar{\bar{y}})$ and $\sigma(\bar{\bar{y}})$ are the predicted mean and standard deviation from GP; $\Phi(.)$ is the cdf; $\phi(.)$ is the pdf; $\xi \geq 0$ is a small value which is set as 0.01.

**Appendix B. Additional Figures**

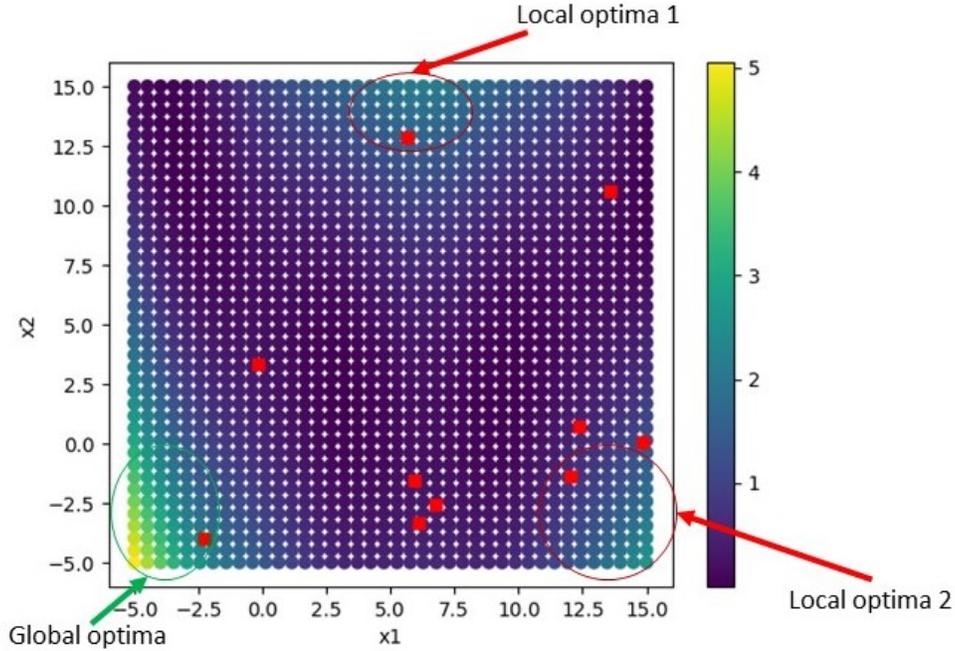

**Figure S1.** Second initialization with 10 starting samples randomly selected (denoted by red dots). Here, at least one sample is located very near to the optimal regions.

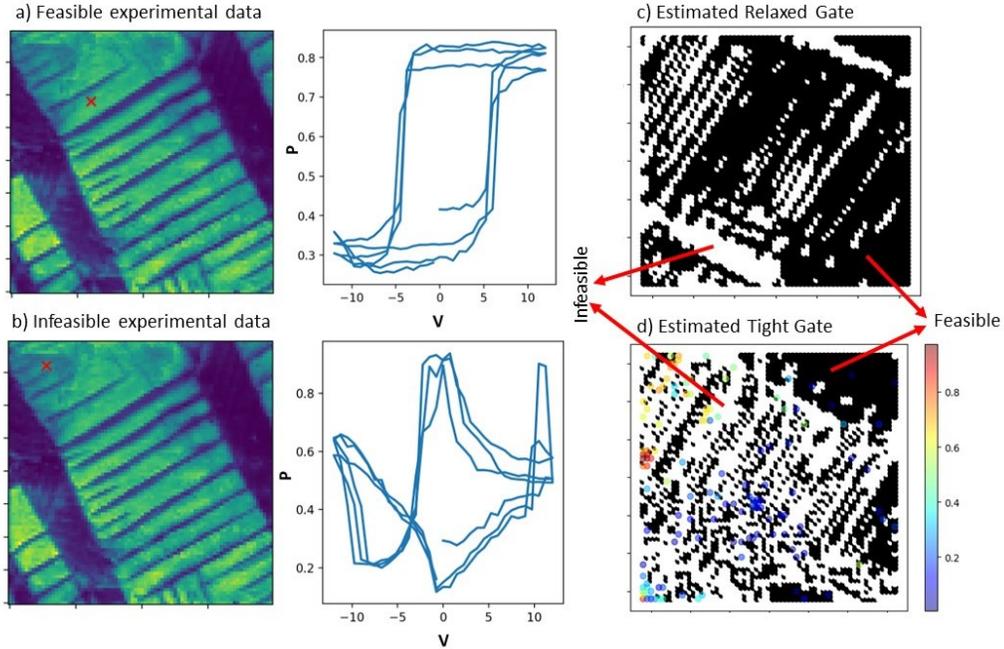

**Figure S2.** Figs (a),(b) show an example of the spectral with positive and negative assessment respectively. These two samples are among the initial 30 samples generated from LHS approach. Fig (c) shows the map of the relaxed gate estimated from the initial data only. Fig (d) shows the map of the tight gate estimated after the exploration cost is exhausted. It is to be noted here that the gate is tightened dynamically as the SANE exploration continues, however, without human intervention. The human assessment is only done for the initial 30 samples. We can see there are some samples selected but located in the estimated infeasible region, those were the part of the feasible region during the respective stages of the iterations. Thus, tightening the constraint also seems to gradually avoid over exploration of a certain region too. However, it can induce the SANE to miss a region of interest (refer to Fig. 6c,d), particularly in the later stage of exploration. Therefore, tuning the penalty factor, $P$ is another alternative to relax the gate.